\documentclass[10pt,twocolumn,letterpaper]{article}

\usepackage{iccv}
\usepackage[misc]{ifsym}  
\usepackage[accsupp]{axessibility}  
\usepackage{amsfonts}
\usepackage{times}
\usepackage{epsfig}
\usepackage{graphicx}
\usepackage{amsmath}
\usepackage{amssymb}
\usepackage[font={small}]{caption}
\usepackage{booktabs}
\usepackage{color}
\usepackage{xcolor}
\usepackage{colortbl}
\usepackage{tikz}
\definecolor{dashedblue}{RGB}{68,114,196}
\definecolor{dashedgreen}{RGB}{169,209,142}
\definecolor{lightyellow}{RGB}{227,223,60}

\newcommand{\matr}[1]{\mathbf{#1}}
\newcommand{\bluedashedline}{\raisebox{2pt}{\tikz{\draw[-,dashedblue,dashed,line width = 1.pt](0,0) -- (5mm,0);}}}
\newcommand{\greendashedline}{\raisebox{2pt}{\tikz{\draw[-,dashedgreen,dashed,line width = 1.pt](0,0) -- (5mm,0);}}}

\usepackage{pifont} 

\newcommand{\xmark}{\ding{55}}%
\usepackage{multirow}

\definecolor{citecolor}{RGB}{65,105,225}
\usepackage[pagebackref=true,breaklinks=true,letterpaper=true,colorlinks,
citecolor=citecolor,bookmarks=false]{hyperref}

\iccvfinalcopy 

\def\iccvPaperID{1580} 

\ificcvfinal\fi

\begin{document}


\title{UnitedHuman: Harnessing Multi-Source Data for\\High-Resolution Human Generation}


\author{
Jianglin Fu\textsuperscript{1*}, \quad Shikai Li\textsuperscript{1*}, \quad Yuming Jiang\textsuperscript{2},\quad Kwan-Yee Lin\textsuperscript{1,3},  \quad Wayne Wu\textsuperscript{1$\dagger$}, \quad Ziwei Liu\textsuperscript{2$\dagger$} \\
$^{1}$ Shanghai AI Laboratory \quad $^{2}$ S-Lab, Nanyang Technological University\quad $^{3}$ CUHK \\
{\tt\small \{fujianglin, lishikai\}@pjlab.org.cn} \\ {\tt\small \{yumingj80, wuwenyan0503\}@gmail.com} \quad {\tt\small ziwei.liu@ntu.edu.sg}\quad {\tt\small junyilin@cuhk.edu.hk}
}


\twocolumn[{%
\renewcommand\twocolumn[1][]{#1}%
\maketitle
\begin{center}
    \centering
    \vspace{-20pt}
    \includegraphics[width=.95\textwidth]{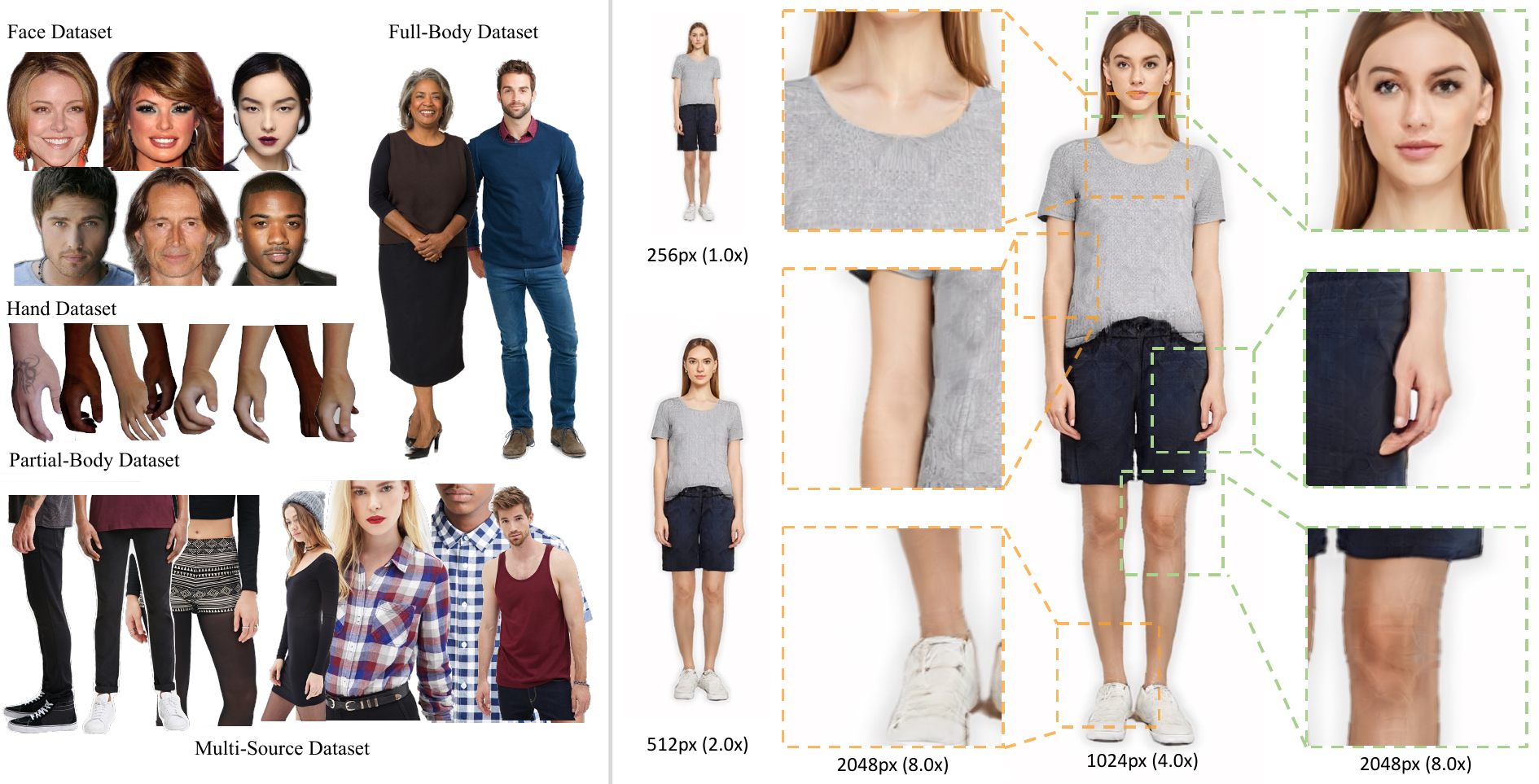}
    \captionof{figure}{\textbf{UnitedHuman.}
    We present our method UnitedHuman that 
    \textbf{left)} integrates multi-source datasets related to human into the full-body image space and 
    \textbf{right)} generates full-body human images at multiple resolutions, and captures high-quality details of the body-parts across different resolutions.
    }
    \label{fig:teaser}
\end{center}
}]

\ificcvfinal\thispagestyle{empty}\fi

\def\thefootnote{*}\footnotetext{Equal contribution.}\def\thefootnote{\arabic{footnote}}
\def\thefootnote{$\dagger$}\footnotetext{Equal advising.}\def\thefootnote{\arabic{footnote}}



\begin{abstract}

Human generation has achieved significant progress. Nonetheless, existing methods still struggle to synthesize specific regions such as faces and hands. We argue that the main reason is rooted in the training data. 
A holistic human dataset inevitably has insufficient and low-resolution information on local parts.
Therefore, we propose to use multi-source datasets with various resolution images to jointly learn a high-resolution human generative model. However, multi-source data inherently \textbf{a)} contains different parts that do not spatially align into a coherent human, and \textbf{b)} comes with different scales. To tackle these challenges, we propose an end-to-end framework, \textbf{UnitedHuman}, that empowers continuous GAN with the ability to effectively utilize multi-source data for high-resolution human generation. Specifically, \textbf{1)} we design a Multi-Source Spatial Transformer that spatially aligns multi-source images to full-body space with a human parametric model.
\textbf{2)} Next, a continuous GAN is proposed with global-structural guidance and CutMix consistency. Patches from different datasets are then sampled and transformed to supervise the training of this scale-invariant generative model. 
Extensive experiments demonstrate that our model jointly learned from multi-source data achieves superior quality than those learned from a holistic dataset. Project page: \url{https://unitedhuman.github.io/}.

\end{abstract}

\section{Introduction}
Human generation tasks have been intensively explored recently, as synthesized photo-realistic human images can benefit various related applications, such as virtual try-on, movie production, etc. Despite the great success of face generation since the emergence of StyleGAN~\cite{stylegan}, early attempts at human generation tasks~\cite{insetgan,styleganhuman,text2human} exhibit limited generative capabilities, especially in producing high-resolution full-body humans.

We argue this is due to intricately articulated human structures and limited training datasets. Specifically, since local parts like hands and faces only occupy a small portion of the entire image, and as a result, they cannot provide the model with sufficient texture information. Unfortunately, to the best of our knowledge, a comprehensive dataset capable of representing highly-detailed visual information on various human body parts is sorely lacking. Also, collecting such a dataset from scratch is time-consuming and labour-intensive.

Despite the scarcity of holistic human datasets, a vast quantity of human partial-body data is accessible to assist scholars in finishing multifarious human-related tasks~\cite{styleganhuman,dart,deepfashion,celeba}. A trivial solution is to supplement the generation process with multiple datasets of human body parts. This simple idea is promising since the existing human-related datasets are expected to enhance details of local body parts, and these multi-source datasets constructed from different groups maintain a high degree of diversity in several aspects, including image scale, illumination, and body part position. Also, datasets of body components offer more ample texture details compared to full-body datasets. Both of these satisfy our needs for human generation and motivate us to utilize multiple human-related datasets to generate high-resolution human bodies. 

To unite these multi-source datasets to push the limit of human generation, we analyze the difficulty of this endeavor and find that the main obstacle is twofold. 1) It is challenging to align disparate body parts into a coherent, realistic human since the scales and locations of the body components in each dataset have different distributions. Merely aligning with 2D keypoints~\cite{openpose} is feasible for rigid objects but suboptimal for hinged human structures, as it disregards depth information as well as the body shape. A reliable alignment mechanism is therefore required to connect these body components. 2) Image resolution varies among multi-source datasets, and this work requires training with these datasets to synthesize results at different resolutions. Multi-scale generation is another necessity that needs to be addressed because GAN models are typically trained on identical-resolution images and can only synthesize images of a fixed resolution.

\begin{figure*}
    \centering
\includegraphics[width=\textwidth]{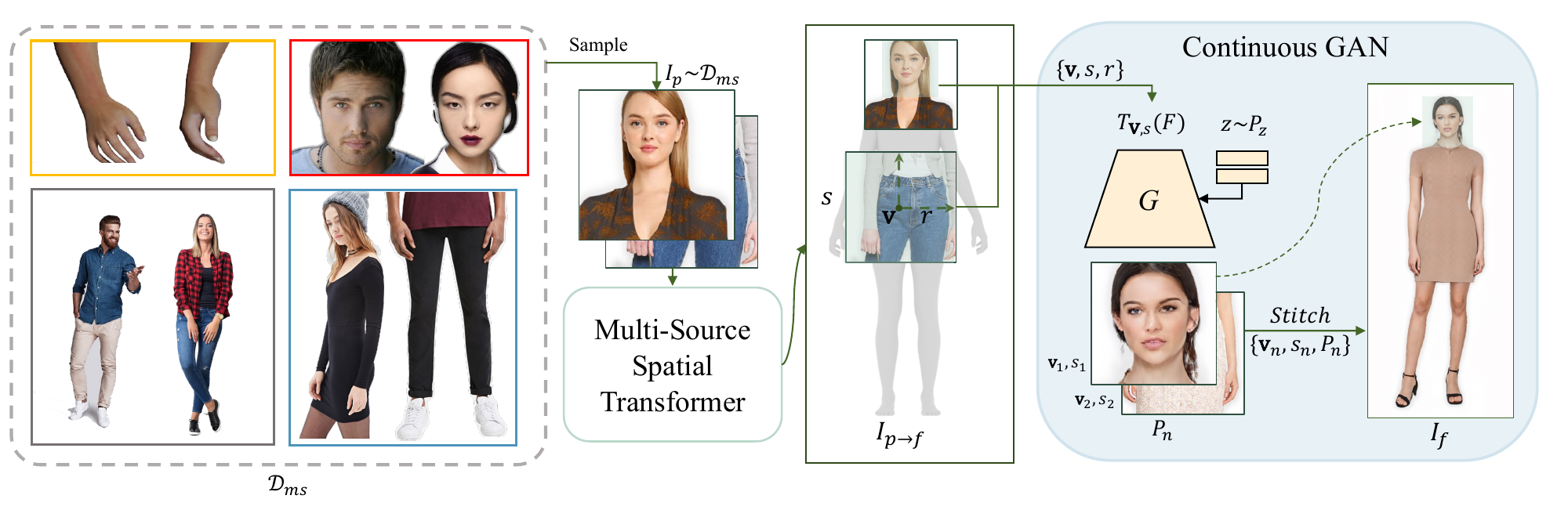}
    \vspace{-7mm}
    \caption{\textbf{Overview of UnitedHuman.} Given the images~$I_p$ from the multi-source datasets~$D_{ms}$, the Multi-Source Spatial Transformer puts the partial-body image into the full-body image space as $I_{p{\rightarrow}f}$ for a unified spatial distribution. With sampling parameters $\textbf{v}, s, r$ and latent code $z$ from the prior distribution, our Continuous GAN generates the patches $P_{n}$ at center $\textbf{v}$ with scale $s$. The patches over the full-body space are stitched to form the high-resolution full-body images~$I_f$.}  
    \label{fig:pipeline}
\end{figure*}

To tackle the above two challenges, we propose \textbf{UnitedHuman}. This end-to-end framework, consisting of \textit{Multi-source Spatial Transformer} module for spatial alignment and \textit{Continuous GAN} module for arbitrary-scale training, leverages multiple datasets to synthesize higher-resolution full-body images. 
Fig.~\ref{fig:pipeline} illustrates our entire working pipeline. Specifically, the\textit{ Multi-Source Spatial Transformer} employs the parametric human model as a prior. This transformation serves to convert the partial-body image into the full-body image space, leading to a unified spatial distribution.
With different sampling parameters in the full-body image space and latent code from the prior distribution, the \textit{Continuous GAN} takes the transformed Fourier feature as input and generates the fixed-resolution patches at corresponding positions and scales. 
Finally, the generated patches over the full-body space are stitched to form high-resolution full-body images. 

Compared to the existing cutting-edge human-GAN models, \textit{UnitedHuman} demonstrates the ability to incorporate human-related datasets from multiple sources to produce high-resolution humans. The humans generated by our model, with zoomed-in details from 256px to 2048px, are shown on the right side of Fig.~\ref{fig:teaser}. During experiments, we demonstrate that by leveraging only $10\%$ of the high-resolution images needed by the SOTA methods, along with the incorporation of various partial datasets from multiple sources, our technique can outperform the existing state-of-the-art outcomes. Furthermore, the model has the potential to scale up to any resolution by introducing additional partial-human datasets. 

In summary, our main contributions are listed: 1) We propose UnitedHuman, the first work to explore the usage of multi-source data for high-resolution human generation tasks. 2) We design \textit{Multi-source Spatial Transformer} to assist in aligning body parts from diverse datasets, based on the articulated human structure. 3) We design the \textit{Continuous GAN} to achieve multi-resolution and scale-invariant training.

\section{Related work}
\label{sec:relatedwork}

\noindent\textbf{Human Image Generation.} 
Scholars in the 2D human generation field commonly adopt GAN models as paradigms: they either generate humans unconditionally~\cite{styleganhuman} or are conditioned on pose/semantic priors~\cite{pose_with_style,spg_net,qian2018pose,tang2020xinggan,zhang2022exploring,casd,patn}, which explicitly alleviates the entanglement between pose and appearance. Other image-based researchers exploit diffusion models to achieve human-associate tasks, such as head swapping~\cite{hsdiffusion} and person image synthesis~\cite{bhunia2022person}. Although the above methods can generate full-body humans, they have certain limitations when it comes to reproducing intricate details of local parts. One recent work, InsetGAN~\cite{insetgan}, proposed a novel method for integrating multiple GAN models to alleviate artifacts in the generated human images. This pipeline iteratively searches and combines latent codes from the pretrained GAN models, and the found latent codes can produce plausible full-body humans. This work demonstrates the possibility of approximating the latent codes of two or more pretrained models from different distributions to each other to generate a coherent full human body. Unlike InsetGAN, we focus on how to leverage different existing datasets to establish an end-to-end framework for high-resolution human generation.

\noindent\textbf{Multi-scale Generation.}  
Requiring models to produce images in multi-resolution or even higher-resolution raises the topic of multi-scale generation. The prevalent approach is to stitch the generated independent patches to construct a full-size image~\cite{anyres,infinitygan,scaleparty}. The leading challenges of this task are generating patches according to specific positions and maintaining global structure across scales. The former is often solved by employing positional encoding~\cite{ms-pe,infinitygan,scaleparty,ms-pie}; consistency loss~\cite{anyres} or global discriminator~\cite{scaleparty} are often employed as a method to overcome the later challenge. Consequently, AnyRes~\cite{anyres} suggests a novel continuous-scale training architecture and exploits datasets with arbitrary resolutions. Although AnyRes can be trained on datasets of arbitrary size, it still requires the training set to have some similarity in distribution. In Sec.~\ref{sec:exp}, we show AnyRes cannot generate plausible full-body humans with local-part datasets, which means it cannot take advantage of partially human datasets. To tackle this issue, we design a unique alignment module targeting human structure.

\noindent\textbf{Human-centric datasets.} 
Human-centric tasks are long-standing and always attract everyone's attention. Although the tasks related to humans are diverse and complicated, the solution to these downstream tasks can greatly improve productivity and contribute to the entire community. Researchers in academia and industry construct large-scale human-related datasets~\cite{dataset_gnr,vitonHD,styleganhuman,dart,dataset_h36m_pami,stylegan,deepfashion,celeba,dataset_interhand,dataset_mead,dataset_lab,dataset_humbi,freihand} through collection or synthesis methods to solve various human-related problems. For example, face generation commonly employs FFHQ~\cite{stylegan} and CelebA~\cite{celeba}, hands synthesis has DART~\cite{dart} and FreiHAND~\cite{freihand}, DeepFashion~\cite{deepfashion} and Viton-HD~\cite{vitonHD} are released for fashion item generation/recommendation, SHHQ~\cite{styleganhuman} is constructed for full-body generation. We select four representative datasets to conduct experiments. This paper first adopts the downsampled SHHQ as a benchmark to provide global information on human bodies, as SHHQ delivers the highest-resolution full-body images. Then, to offer high-definition textures, the high-resolution versions of DeepFashion and SHHQ are also included. DART and CelebA are added to enhance the local details of hands and faces respectively.

\section{UnitedHuman}

Our aim is to generate full-body images from multi-source datasets of varying distributions and resolutions. As shown in Fig.~\ref{fig:pipeline}, Multi-Source Spatial Transformer aligns the spatial distributions among the multi-source datasets within the full-body image space~(Sec.~\ref{sec:spatial_trans}). Given the spatial and scale parameters in the image space, the proposed Continous GAN will generate the patches to form the full-body images with global-structural guidance and CutMix consistency~(Sec.~\ref{sec:continuousgan}). 

\subsection{Multi-Source Spatial Transformer}
\label{sec:spatial_trans}
Previous works~\cite{stylegan,stylegan2} require full-content images with an aligned spatial distribution such as FFHQ and AFHQ. While the SHHQ dataset~\cite{styleganhuman} provides full-body images for human generation, it can only reach a resolution of 1024, which is insufficient for generating specific regions such as faces and hands.
Therefore, we set our sights on vast human partial-body data with various spatial distributions but provide high-resolution parts. To unite these datasets for human generation, we propose the Multi-Source Spatial Transformer that transforms different partial-human images into a defined full-body image space using the parametric human model as prior.

We first define the full-body image space as a bounded region that represents the spatial distribution of the human from the well-aligned full-body dataset. Note that the partial-body images can also be placed in this image space with appropriate transformation. 
To represent both the full-body and partial-body human in a unified manner, we choose the parametric model SMPL~\cite{smpl} as the geometry prior. 
Specifically, given a full-body dataset, we first employ a model-based human reconstruction method~\cite{pare} to estimate the body parameters of pose ${\theta}$, shape ${\beta}$, and a weak-perspective camera model $\alpha$ with scale and translation parameters. 
As for the issue of scale ambiguity in monocular images, we keep the camera intrinsics fixed and regard the estimated body heights as approximate, heavily constrained by the shape prior. Finally, our full-body image space $S$ can be simply defined by rendering the bounded region with the camera parameters~$\bar{\alpha}$.

Proper camera parameters of partial images are also required for placing the images within the full-body image space. However, inaccurately predicted parameters result from the lack of suitable training data under the partially-observed setting for human mesh recovery. To address this, we follow the regression-optimization hybrid manner~\cite{smplify_x_partial} with the assistance of a full-body dataset. Specifically, we decompose partial images into visible and invisible parts by the predicted 2D keypoints~\cite{openpose}. After the initial regression on partial images, we optimize the pose parameters of visible parts by minimizing the error between projected 3D keypoints and estimated 2D keypoints. For unseen parts, we use a variational autoencoder~(VAE) trained on pose parameters of the full-body dataset as the pose prior for optimization. We also propose an additional orientation regularization to regularize the pitch value of the global orientation $o$ to prevent the problem of depth ambiguity. Although the optimization focuses on pose parameters, the weak-perspective camera parameters can also be more reasonable. The overall loss of optimization is then defined as
\begin{equation}\label{e_kopt}
    \begin{split}
    E({\alpha}_{opt}) 
    &= L_{vis} + L_{invis} + L_{reg}\\
    &= \sum^{\text{vis}}_{i}{\mathcal{L}(J_{i}, J^{\prime}_{i})} + \sum^{\text{invis}}_{i}{ \mathcal{L}(\theta_{i}, \hat{\theta}_{i} )} +\mathcal{L}(o, \hat{o})
    \end{split}
\end{equation}
where the $J$ and $J^\prime$ denote the projected keypoints with camera ${\alpha}_{opt}$ and estimated 2D keypoints respectively while the $\theta$ and the $\hat{\theta}$ denote the axis-angle of joints from the predicted SMPL and VAE. The mean squared error (MSE) loss is used in Eq.~\ref{e_kopt}. Finally, the partial images $I_p$ are transformed into the full-body image space by
\begin{equation}
    I_{p\rightarrow{f}} = H( {\alpha}, \bar{{\alpha}}) \times I_p
\end{equation} 
where $H$ is the matrix that calculated by the optimized camera ${\alpha}_{opt}$ and average camera $\bar{{\alpha}}$ of the full-body dataset.
\begin{figure}
    \centering
\includegraphics[width=1\linewidth]{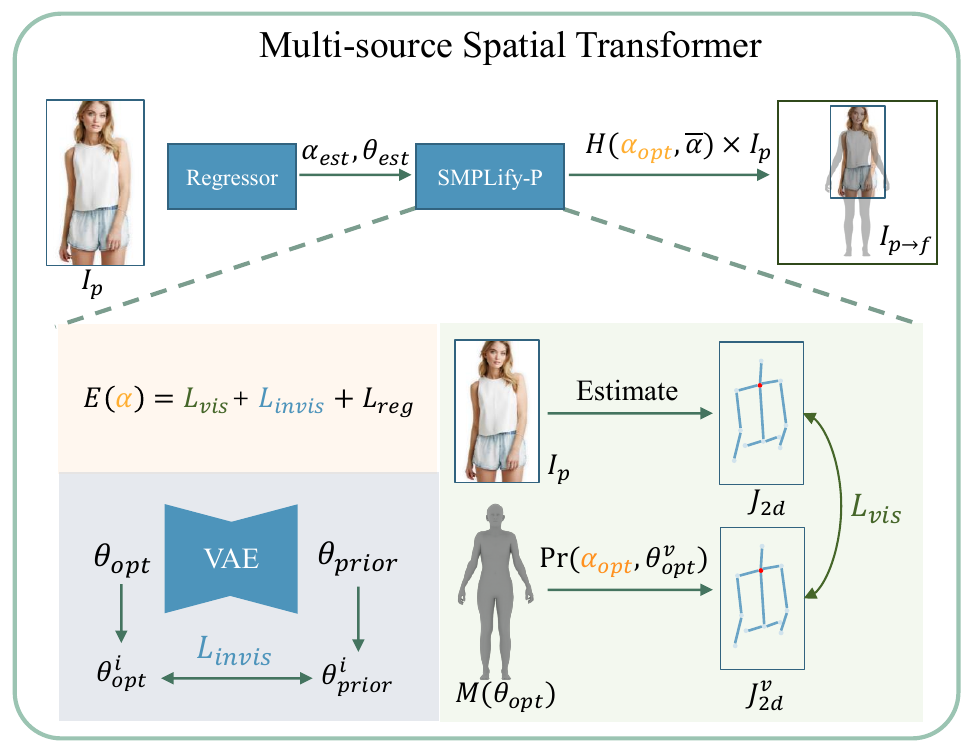}
    \caption{\textbf{Multi-source Spatial Transformer.} Given a image $I_p$, we first predict the camera ${\alpha}_{est}$ and pose $\theta_{est}$ of SMPL. Optimized by SMPLify-P on both visible and invisible parts, the camera ${\alpha}_{opt}$ is used to calculate the matrix $H$ that transforms the patch into the full-body image space.}
    \label{fig:MST}
\end{figure}
\subsection{Continous GAN}  
\label{sec:continuousgan}

\noindent\textbf{Preliminary.}~Generative adversarial network~\cite{goodfellowGAN} is proposed to synthesize images with a fixed resolution. Continuous-resolution generative models~\cite{anyres,scaleparty} extend the approach to arbitrary-scale training by transforming the input embedding and applying supervisions between different scales. 
The generation of the image can be viewed as sampling values at discrete positions within a bounded region of continuous image space ranging from $[0,0]$ to $[1,1]$. In the sampling process, we define the center $\textbf{v}$ as the coordinate center of the sampled patch while scale $s$ and resolution $r$ represent sampling frequency and the number of sampling points, respectively. 
For image sampling in the image space, the continuous-resolution generator $G$ synthesizes the patch’s pixel values at the sampled coordinates as:
\begin{equation}
\begin{split} \label{eq:prel}
G(z, \text{emb}, \textbf{v}, s ,r) 
&= G(T_{\textbf{v},s,r}(\text{emb}), z) \\
&= T_{\textbf{v},s,r}(G(\text{emb}, z))
\end{split}    
\end{equation}
where $T$ is the transform function based on the sampling parameters $(\textbf{v}, s, r)$ and $z$ is the latent code sampled from the prior distribution. As seen in Equation~\ref{eq:prel}, the continuous-resolution generator also owns the following properties: spatial equivariance and scale consistency. 

\noindent\textbf{Continuous Network.}~
We constructed our network based on StyleGAN3-T~\cite{stylegan3} with a pixel-wise discriminator to fully utilize the multi-source dataset. Although the architecture of the generator is proposed for anti-aliasing and spatial equalvariance,  it is able to generate the images at a slightly larger scale by directly modifying the sampling frequency of the Fourier feature. With additional training and supervision at different scales, we could utilize the transformation matrix $T$ to control the image generation in the full-body image space.  Specifically, given the sampling parameters $(\textbf{v}, s)$ with a fixed resolution $r$, the transformation matrix $T$ of input Fourier function is as follows:
\begin{equation}
T_{\textbf{v}, s} = \begin{pmatrix}
\frac{1}{s} & 0 & \matr{v_x}-0.5 \\
0 & \frac{1}{s} & \matr{v_y}-0.5 \\
0 & 0 & 1
\end{pmatrix}
\end{equation}
where $T$ is an identity matrix that samples the full-body image in the continuous space when $(\textbf{v}, s) = ([0.5, 0.5], 1)$. As for image patch generation with large scale $s$ or different position $\textbf{v}$, the two parameters of the Fourier feature, frequency $f$ and phase $p$, can be transformed by $T$ to represent the transformed sampling patch in the image space. Notably, the Nyquist–Shannon sampling theorem~\cite{Shannon1949} is still satisfied when ${s > 1}$ for anti-aliasing generation~\cite{stylegan3}. The Fourier embedding $F(f,p)$ will be fed to the generator $G$ to synthesize the image~$x$ as follows:
\begin{equation}
x = G(T_{\textbf{v},s}(F)), z)
\end{equation}
where $z$ denotes the latent code from the prior distribution. A pixel-wise loss between generated patches at different scales is applied to achieve scale consistency.

So far our network supports the scale-invariant training on a holistic multi-resolution human dataset with a typical discriminator. However, the images from the partial-body datasets cannot occupy the entire continuous image space, resulting in partial-body image generation. One trivial solution is to sample the image patches inside the subregion of image space, where these patches can only be seen at specific ranges of position and scale. To overcome this issue, we apply the pixel-wise discriminator with a proposed CutMix consistency for multi-source dataset training. Following \cite{unetD}, we alter the architecture of $D$ to an encoder-decoder network by applying the upsampling blocks and skip connections, resulting in performing pixel-wise classification of input image. 
As for partial-body image patches, we apply the CutMix operation by mixing the real patches and the generated ones by the mask of the subregion as:
\begin{equation}
\text{CutMix}(x, \hat{y}, M)= x \odot (1-M) + \hat{y} \odot M
\end{equation}
where $M$ is the mask inside the subregion and $\odot$ is the element-wise production. Therefore the partial-body data can be sent to the discriminator at all scales during training, bringing a more powerful discriminator.

\begin{figure}
    \centering
\includegraphics[width=1\linewidth]{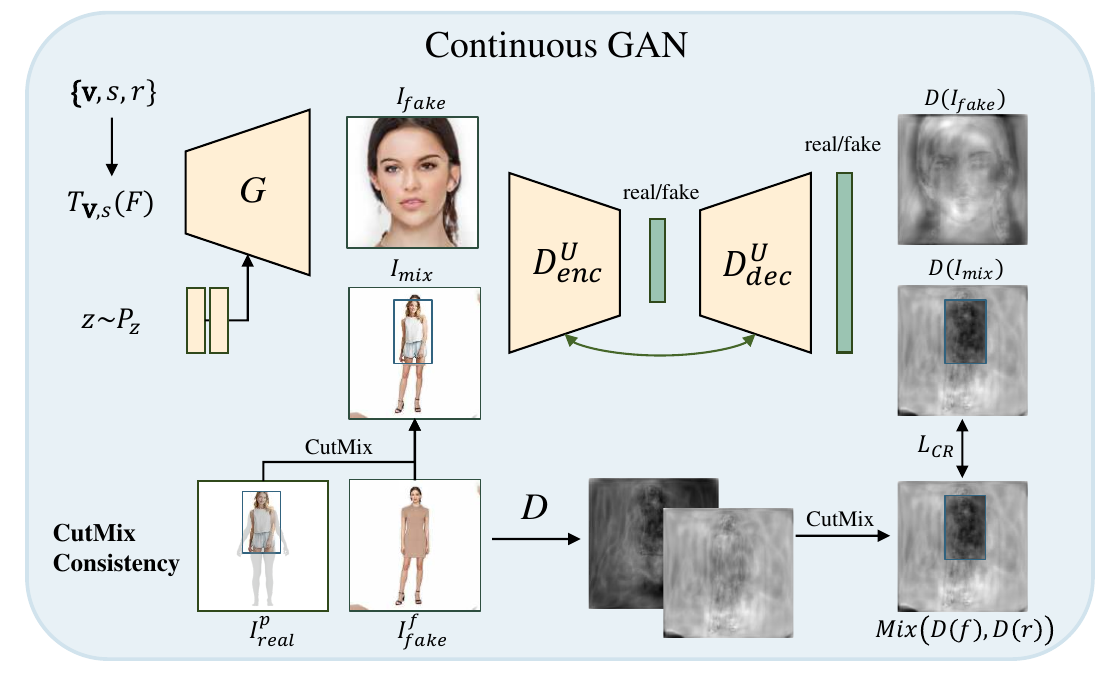}
    \caption{\textbf{Continous GAN}. Given sampling parameters ${(\text{v}, s)}$, the transformed Fourier feature $T_{\text{v}, s}(F)$ is used to generate patches with latent code $z$. We use the U-Net discriminator for global and pixel-wise adversarial training. The proposed CutMix consistency makes the partial-body images $I_{real}^{p}$ trained at all scales.
    }
    \label{fig:ContinousGAN}
\end{figure}

\noindent\textbf{Multi-source dataset training.}~We implement the training among multi-source datasets in a two-stage manner. During the \textit{Stage 1}, we train the basic model on a low-resolution full-body dataset as global structural guidance. We apply both global and pixel-wise non-saturating GAN loss with R1 regularization for \textit{Stage 1} as:
\begin{equation}
    \begin{split}
    \mathcal{L}_{adv,D} & = \mathbf{E}_{\boldsymbol{z}\sim P_z} [f(D(G(T_{\textbf{v},s}(F)), z)) ] \\ & + \mathbf{E}_{\boldsymbol{I}\sim P_{data}} [f(D(\boldsymbol{I})) + \lambda ||\nabla_{\boldsymbol{I}}D(\boldsymbol{I})||^2_2] \\
\mathcal{L}_{adv,G} & = \mathbf{E}_{\boldsymbol{z}\sim P_z}[f(-D(G(T_{\textbf{v},s}(F)), z)) ]
    \end{split}
\end{equation}
where $f(x)=-log(1+e^{-x})$ and $T_{\textbf{v},s}$ is an identity matrix. 

In \textit{Stage 2}, we use the Multi-source Spatial Transformer to sample the image patches from multi-source high-resolution datasets as real images. 
For training efficiency, we also apply the keypoint-based sampling since the human body occupies the horizontal center region mostly. 
As for scale-invariant training schema, the pixel-wise loss between the transformed patches and the full-body image generated by the basic model is also applied for the same latent code $z$  as follows:
\begin{equation}
    \mathcal{L}_{pixel} = (\mathcal{L}_{lpips}(H(x), I_p) + \mathcal{L}_{1}(H(x), I_p)) \odot M_{T}
\end{equation}
where $H$ is another operation that transforms the generated patches to another scale and $M_T$ is the corresponding mask. 
To achieve consistent training on partial images, we utilize the CutMix consistency by regularizing the decoder outputs of mixed image and the mixed of decoder outputs of real partial images and fake full images. The regularization is: 
\begin{equation}
    \mathcal{L}_{cr} = D_{d}(\text{Mix}(x, I_p)) - \text{Mix}(D_{d}(x), D_{d}(I_p))
\end{equation}
where $D_d$ denotes the decoder of pixel-wise discriminator.
Therefore, the overall loss of generator in \textit{Stage 2} can be formulated as follows:
\begin{equation}
\begin{split}
    \mathcal{L}_{G} & = \mathcal{L}_{adv,G} 
    + \lambda_{p}\mathcal{L}_{pixel} \\
    \mathcal{L}_{D} & = \mathcal{L}_{adv,D} + \lambda_{cr}\mathcal{L}_{cr}
\end{split}
\end{equation}
where $\lambda_{p}$ and $\lambda_{cr}$ are $5.0$ and $1.0$ respectively.
\section{Experiments}\label{sec:exp}

\begin{figure*}[htbp]
    \centering
    \includegraphics[width=\linewidth]{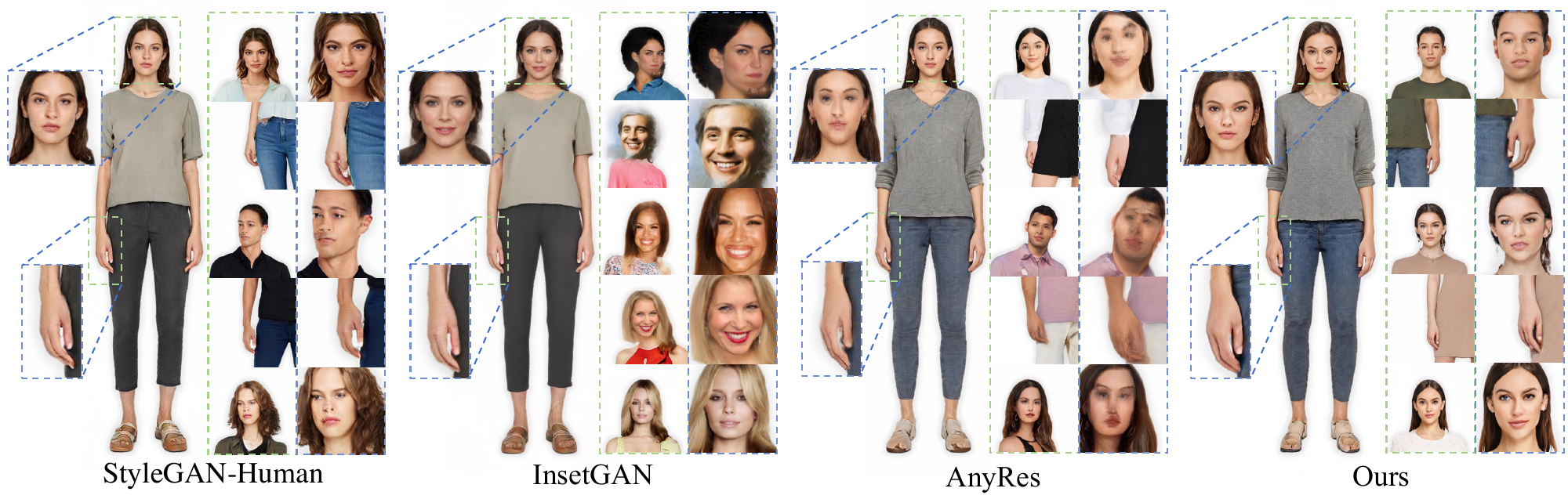}
    \caption{\textbf{Results of baseline comparison}: StyleGAN-Human~\cite{styleganhuman}, InsetGAN~\cite{insetgan}, AnyRes~\cite{anyres}, and UnitedHuman. We exhibit the full-body human images generated from each experiment at a resolution of $1024$ (\protect\greendashedline), as well as the face and hand patches cut from the $2048$px images (\protect\bluedashedline). }%
    \label{fig:baseline_compare}
\end{figure*}

\begin{figure*}[htbp]
    \centering
    \includegraphics[width=0.8\linewidth]{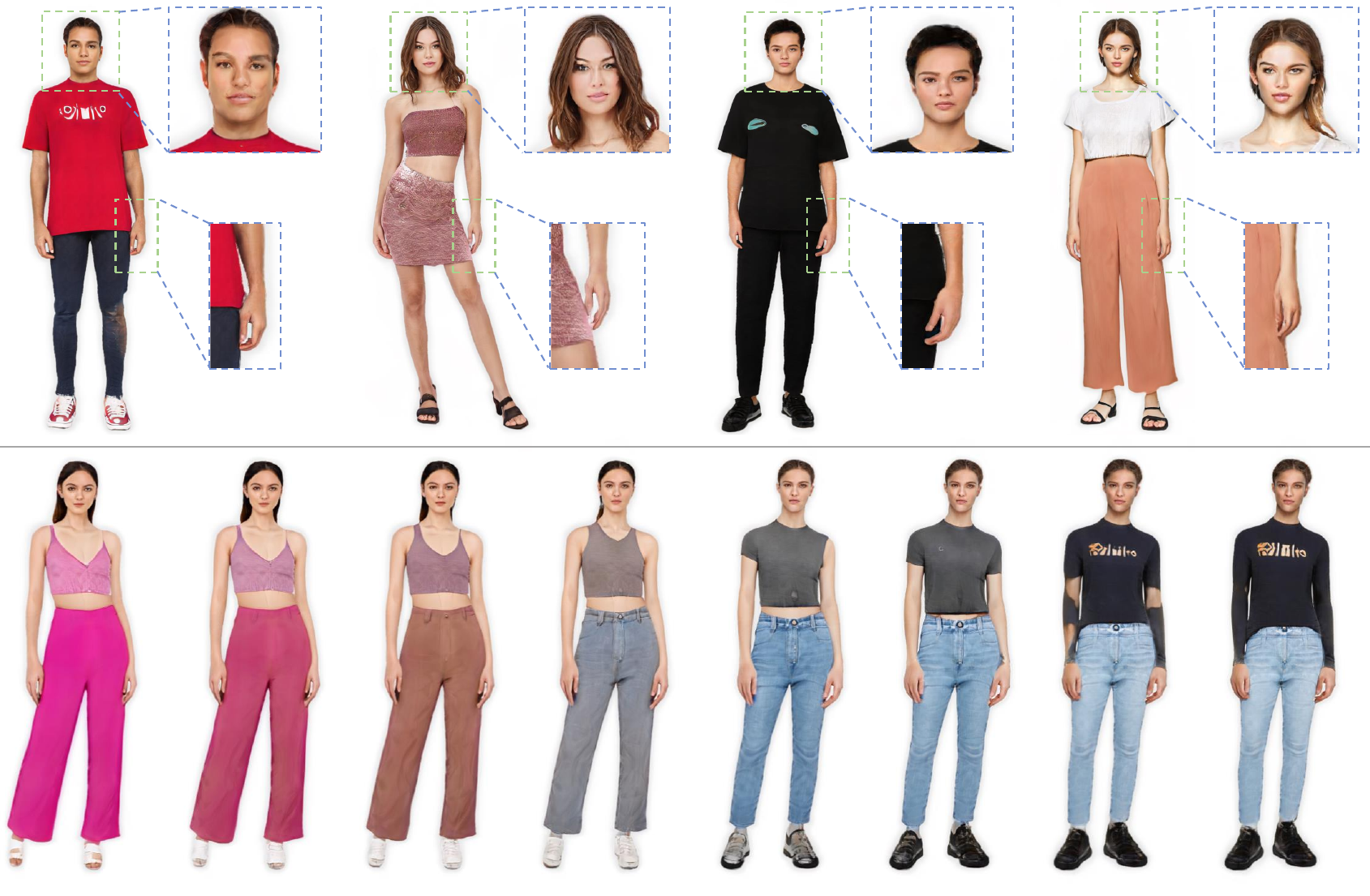}
    \caption{More results. The top row illustrates more UnitedHuman results in $1024$px (\protect\greendashedline) and $2048$px (\protect\bluedashedline).
    An example of interpolation between latent codes is given in the bottom row.}
    \label{fig:more_results}
\end{figure*}

\begin{table*}[htbp]
    \centering
    \begin{tabular}{l|c|c|c|c|c|c|c|c|c|c|c}
    \toprule 
    & \multicolumn{7}{c|}{kFID} &\multirow{2}{*}{$\overline{\text{kFID}}$} & \multirow{2}{*}{pFID} & \multirow{2}{*}{Precsion} & \multirow{2}{*}{Recall} \\ 
    \cline{2-8}
             & Face & Neck & Shoulder & Elbow & Hand & Hip & Knee &  & & &\\ \toprule 
    SG-Human~\cite{styleganhuman} & 22.21 & 21.33 & 19.77 & \textbf{19.42} & 18.47 & 20.01 & \textbf{20.56} & 20.25 & 18.96 & 0.71 & \textbf{0.65}\\
    InsetGAN~\cite{insetgan} &40.84 &37.26 &33.60 &31.99 &25.44 &29.90 &28.49 & 32.50 &27.22 & 0.70 & 0.39 \\
    AnyRes~\cite{anyres} & 33.26 &31.20 &28.55 &33.57 &33.02 &37.97 &34.26 &33.12 &30.49 & 0.67 & 0.48\\
    Ours  & \textbf{21.49} & \textbf{19.48} & \textbf{17.88} & 19.66 & \textbf{17.80} & \textbf{19.62} &  20.96 & \textbf{19.56} & \textbf{18.94} & \textbf{0.74} & 0.61\\
    \bottomrule 
    \end{tabular}
    \caption{\textbf{Quantitative results for baseline comparison.} $\overline{\text{kFID}}$ denotes the average value derived from all kFIDs. All the metrics are measured in the images at $2048\times2048$ pixels. Our method delivers the overall best results, though seeing only 1/10 high-resolution full-body data compared to StyleGAN-Human~\cite{styleganhuman} and InsetGAN~\cite{insetgan}.
    }
    \label{tab:quant_ablation_baseline}
\end{table*}

\begin{table*}[htbp]
    \centering
    \begin{tabular}{l|c|c|c|c|c|c}
    \toprule 
    & LR\textsuperscript{256} & HR\textsuperscript{1024} & \multicolumn{4}{c}{SR\textsuperscript{2048}} \\
    \cline{2-7}
      & SHHQ~\cite{styleganhuman} & SHHQ & CelebA
      \textsuperscript{$256$}~\cite{celeba} & $\text{DF}_{p}$~\cite{deepfashion}& DART~\cite{dart} & SHHQ \\ 
     \toprule %
    SG-Human~\cite{styleganhuman} 
    & \xmark & $100K$& \xmark &\xmark&\xmark&\xmark\\
    InsetGAN~\cite{insetgan}
    & \xmark &$100K$& $30K$ & \xmark &\xmark&\xmark\\
    AnyRes~\cite{anyres} 
    & $100K$ &$10K$&\xmark&\xmark&\xmark&\xmark \\
    Ours &$100K$ &$10K$ & $10K$ &$10K$ & $7K$ & $5K$ \\
    \bottomrule 
    \end{tabular}
    \caption{\textbf{Data composition} seen by baseline models and the proposed UnitedHuman. The datasets can be divided into three categories based on resolution: $256$px (LR), $1024$px (HR), and $2048$px (SR). We treat CelebA in $256$ pixels as an SR dataset because the faces in CelebA are of similar size to the faces in the full-body images at $2048$px. $\text{DF}_{p}$ refers to the partial human images from DeepFashion dataset. SHHQ\textsuperscript{SR} is obtained by up-sampling SHHQ\textsuperscript{HR} and taking only the lower half of the images. It is used to supplement the human lower-body dataset. We show that UnitedHuman can digest datasets from different data sources in diverse resolutions.}
    \label{tab:datacomposition}
\end{table*}

\subsection{Experimental Setup}
The whole work is trained successively through two phases as discussed before. \textit{Stage 1} is trained with full-body images in a resolution of $256\times256$ pixels, whereas \textit{Stage 2} employs \textit{Stage 1} model as the teacher model and further refines body-part details with multi-source datasets in higher-resolutions. 

\noindent\textbf{Datasets.} 
In this work, experiments are conducted on the following human-centric datasets: SHHQ~\cite{styleganhuman}, DeepFashion~\cite{deepfashion}, CelebA~\cite{celeba}, and DART~\cite{dart} that is rendered by Blender's Eevee~\cite{blender}. Refer to Tab.~\ref{tab:datacomposition} for more details.

\noindent\textbf{Evaluation Metrics.} 
Fr\'echet Inception Distance (FID) is one common indicator for accessing GAN models. However, studies~\cite{insetgan,styleganhuman} argues that FID is more sensitive to diversity in data distribution but struggles to accurately evaluate the visual quality of a human body. In this paper, we jointly utilize two variants of FID to better quantify the model performance. 
1) \textbf{patch-FID (pFID)} is proposed in AnyRes GAN~\cite{anyres} to better assess the local textures in high-resolution images. It randomly samples image patches from high-resolution datasets and stores each transformation matrix. The stored matrixes are then injected into the generator to produce image patches of correlated scales and positions. 2) \textbf{keypoint-FID (kFID)} can be treated as a special case of pFID. It calculates pFID patches around each body keypoint at a specific scale. This kFID is more closely connected to human keypoints, and is able to more clearly reveal the intricate texture representation surrounding the joint points with various degrees of freedom.
To be precise, we employ the sampling parameters $\textbf{v}$ and $s$ to extract patches from both the generated full-body images and the training dataset.  In particular, we fix the scale $s$ at $8$ to ensure the images are generated with a resolution of $2048$ pixels. Then we randomly select a central point $\textbf{v}$ ranging from $[0,0]$ to $[1,1]$ and crop patches around this center point $\textbf{v}$.

\noindent\textbf{Comparison Methods.} To demonstrate the efficacy of our approach for generating plausible humans in high resolution by uniting multi-source datasets, we compare UnitedHuman with three baseline models: StyleGAN-Human~\cite{styleganhuman}, InsetGAN~\cite{insetgan} and AnyRes GAN~\cite{anyres}. Tab.~\ref{tab:datacomposition} depicts detailed data composition that is employed to train each model.
Since StyleGAN-Human does not support training images with multiple resolutions and different body parts, we train it with $100K$ SHHQ\textsuperscript{\textbf{HR}} images in $1024\times1024$ pixels. InsetGAN serves as a multi-GAN optimization technique for merging face and body images created by separate GAN models. In our study, we adopt the public pre-trained CelebA~\cite{celeba} model with a resolution of $256\times256$ and the StyleGAN-Human model we previously trained as the two fundamental models for InsetGAN.  Lastly, AnyRes allows continuous-scale training using a two-stage approach but lacks the ability to generate coherent individuals by combining body parts from multi-source datasets. As a result, we train its first stage using $100K$ SHHQ\textsuperscript{\textbf{LR}}, and then employ $10K$ SHHQ\textsuperscript{\textbf{HR}} for the second stage of training. 

\noindent\textbf{Fairness of Comparison.}
The cross-method comparison is conducted fairly and impartially. To ensure a fair comparison, all the compared SOTAs are fully trained till they converged using the configurations of their reported best models. The main difference is the data composition (Tab.~\ref{tab:datacomposition}). Comprehensive training information for all the compared methods, encompassing training parameters and the time it takes for inference, can be found in Appendix~\ref{exp}.

\subsection{Main Results}
\label{sec:main_results}
Fig.~\ref{fig:baseline_compare} depicts the results of the visual comparison of our model against the three baseline methods.
In the figure, we show the full-body humans ($1024\times1024$ pixels) and the crops of the face and hand (obtained from $1024$px and $2048$px full-body humans), respectively. In addition, the full-body images in the figure are derived from the mean latent code of each model. As described in the above section, StyleGAN-Human is trained with a single holistic dataset, and it only supports the generation of fixed-size human images; therefore, the cropped faces and hands are up-scaled from the images in $1024\times1024$ pixels using bi-cubic interpolation. The same approach is used for InsetGAN to obtain image patches from $2048$px. In contrast, AnyRes and our model are able to synthesize images with varied resolutions, and we show the patches cropped from the generated $2048$px images. The figure indicates that the StyleGAN-Human model, which is trained on $100K$ high-resolution full-body images, is capable of producing plausible human images at $1024$ pixels. However, attempting to up-sample these images to higher resolutions brings in artifacts. InsetGAN delivers less desirable results when the distribution of face and body is far apart. The results of AnyRes show that even trained on multi-scale data from the same distribution, it cannot effectively map articulated human body structures to different resolutions. Besides, UnitedHuman generates human images in high resolution with greater details while maintaining the overall human structure. More results generated by UnitedHuman are illustrated in Fig.~\ref{fig:more_results}.

To quantitatively measure the advancement, Tab.~\ref{tab:quant_ablation_baseline} records the numerical numbers of different evaluation metrics. We utilize the SHHQ\textsuperscript{HR} dataset as the real data in FID calculations to guarantee equitable comparisons, since all four methods have seen this dataset. In addition, we do not have a ground-truth dataset that allows us to evaluate model performance in $2048$ pixels. Therefore, we down-sampled all $2048$-pixel images back to $1024$px before commencing evaluation. As demonstrated in the table, UnitedHuman outperforms InsetGAN and AnyRes models in terms of kFID at each body keypoints as well as pFID. Comparing our approach to StyleGAN-Human, we are not surprised that the evaluation outcomes of UnitedHuman do not show a noteworthy advantage in terms of kFID and pFID. We argue that the reason for this phenomenon is that StyleGAN-Human is trained on the single, holistic SHHQ dataset only, the model encounters a simpler and more homogeneous data distribution. Furthermore, the model is assessed using the same SHHQ dataset, which possesses the inherent benefit of computing FIDs. For UnitedHuman, incorporating diverse datasets in the training process presents a more significant difficulty. Besides, we only leverage $10K$ SHHQ\textsuperscript{HR} images along with other partial datasets to train the model. In this scenario, we are still able to achieve comparable and slightly better results than StyleGAN-Human.

In addition, our model achieves an FID score of $15.37$, slightly higher than StyleGAN-Human's $13.81$. Once again, we posit that this difference can be ascribed to StyleGAN-Human being trained on a large-scale holistic dataset. Moreover, it is crucial to note that FID itself may not effectively capture the perceptual quality. We further evaluate the four models with the improved precision and recall metric~\cite{pr}, as shown in Tab~\ref{tab:quant_ablation_baseline}. 
As depicted in the table, UnitedHuman achieves the best precision but obtains marginally lower recall. According to the previous study~\cite{pr}, recall quantifies the fraction of the training data manifold covered by the generator, and we employ a subset of the training dataset (SHHQ\textsuperscript{HR}) used by StyleGAN-Human as our ground truth dataset. Consequently, the StyleGAN-Human model possesses an advantage in recognizing a greater number of true positive instances. On the other hand, our model, having encountered diverse datasets, has generated a data distribution that significantly deviates from the ground truth. This discrepancy has led to some of the generated images being considered negative cases during computation.

In sum, the above results reveal the effectiveness of our model in uniting multi-source datasets with various distributions and improving the details of the generated images. It also indicates that UnitedHuman can synthesize a decent full-body human utilizing a limited amount of high-resolution partial-body datasets.

\subsection{Ablation Study}

\begin{table*}[ht]
    \centering
    \begin{tabular}{l|c|c|c|c|c|c|c|c|c}
    \toprule 
    & \multicolumn{7}{c|}{kFID} &\multirow{2}{*}{$\overline{\text{kFID}}$}& \multirow{2}{*}{pFID}\\
    \cline{2-8} 
    & Face & Neck & Shoulder & Elbow & Hand & Hip & Knee &  & \\ \toprule
    \multicolumn{10}{c}{\textit{Ablation on Datasets}} \\ \midrule

    SHHQ\textsuperscript{\textbf{HR}} &
    25.56 &24.94 &22.28 &25.24 &25.07 &28.15 &27.13 & 25.48 &25.58 \\ 
    + SHHQ\textsuperscript{\textbf{SR}} +$\text{DF}_{p}$ &
    23.43 &22.02 &20.14 &21.75 &19.89 &22.03 &24.61 & 21.98 &21.87 \\  
    + CelebA & 
    21.57 &20.28 &18.99 &21.87 &20.60 &22.86 &24.32 & 21.50 &21.12 \\ 
    + DART (Ours) 
        &\textbf{21.49} &\textbf{19.48} &\textbf{17.88} &\textbf{19.66} &\textbf{17.80 }&\textbf{19.62} &\textbf{20.96} & \textbf{19.56} &\textbf{18.94} \\
    \midrule 
    \multicolumn{10}{c}{\textit{Ablation on Alignment}} \\ \midrule
    Keypoint&
    22.56&21.89&19.77&22.52&20.46&22.91&27.04&22.45 &22.88 \\ 
    Pose-mapping &51.24&22.49&27.01&24.41&19.47&21.54&28.55&27.81 &25.32\\ 
    SMPL (Ours) 
        &\textbf{21.49} &\textbf{19.48} &\textbf{17.88} &\textbf{19.66} &\textbf{17.80 }&\textbf{19.62} &\textbf{20.96} &\textbf{19.56} &\textbf{18.94} \\ 
   
        \bottomrule
    \end{tabular}
    \caption{\textbf{Ablation on datasets and alignment.} All the metrics are measured in the images at $2048\times2048$ pixels.} 
    \label{tab:quant_ablation}
\end{table*}

\noindent\textbf{Ablation on Dataset. } 
We begin with $10K$ SHHQ\textsuperscript{\textbf{HR}} and progressively introduce additional datasets to train UnitedHuman in order to investigate the influence of data on the Multi-source Spatial Transformer. All the experiments, with the exception of data compositions, share the same \textit{Stage 1} teacher model and training hyper-parameters. 
The results are shown in the first section of Tab.~\ref{tab:quant_ablation}.
Start from SHHQ\textsuperscript{HR}, we add $5K$ SHHQ\textsuperscript{SR} and $10K$ DeepFashion data into the model. As mentioned in Tab.~\ref{tab:datacomposition}, the partial images from DeepFashion are mainly focused on the upper-body, while SHHQ\textsuperscript{SR} supplements the lower-body. We observe a considerable enhancement in kFID and pFID after adding these two datasets. Subsequently, the inclusion of CelebA in the pipeline results in a reduction of approximately 1.5 points in kFID around the face, neck and shoulder, while causing an average increase of 0.5 points in other regions. Next, the introduction of the DART dataset leads to a notable enhancement in the kFID around hands. This ablation reflects that the benefits of incorporating body-part datasets outweigh the negative impact.

\noindent\textbf{Ablation on Alignment. }
We conduct experiments to probe the impact of alignment strategy (see the second section of Tab.~\ref{tab:quant_ablation}). When we align human poses, we preemptively take the human images generated from the mean latent code of the teacher model as a reference and compute keypoints by OpenPose~\cite{openpose}, which we denote as \textit{apose}. The ''Keypoint" experiments simply sample the patches of the associated keypoints from the real dataset according to \textit{apose} and use the same transformation matrix to produce the corresponding patches. This naive approach ignores that not all poses of the generated images during training are in line with \textit{apose}. A small offset in pose can cause the training data pairs to be misaligned.
The second ''Pose-mapping'' method trains an auxiliary MLP network to enhance the accuracy when querying the keypoint locations in the generated images during the sampling process. This approach, however, only improves local alignment precision but still cannot cope with the articulated human body structure that contains multiple postures and body shapes. 
Compared to the above two experiments, UnitedHuman incorporated with SMPL achieves better alignment results, as shown in Tab.~\ref{tab:quant_ablation}.

\noindent\textbf{Ablation on Loss. }
Studies are performed to investigate how different loss components affect the outcome. Initially, the full set of the multi-source datasets is trained on StyleGAN3-T with standard adversarial loss. Following that, pixel loss is applied to supervise the global human structure, leading to a substantial reduction in the mean kFID score by $20.7$ points and a corresponding drop in pFID by $17.96$ points.
Afterwards, the pixel-wise discriminator with CutMix consistency loss is added and both the kFID and pFID are dropped to $19.56$ and $18.94$, respectively. This set of experiments demonstrates that low-resolution full-body images can effectively provide structural guidance to the generation of high-resolution humans. Additionally, the use of a pixel-wise discriminator with CutMix consistency allows for the partial-body data to be distributed across different scales and further improves the model performance.
\section{Discussion}

\noindent
\textbf{Limitation and Future Work.}
As we use StyleGAN3, we argue that the underlying architecture may have limitations in representing high-frequency information. This becomes particularly evident when progressively enlarging the image. This limitation is consistent with observations from prior works~\cite{anyres,qiu2022stylefacev}. Furthermore, we've identified that the StyleGAN3 architecture leads to the emergence of a circular grid-like moire pattern. We will further investigate how to alleviate this issue in the future.

We also conduct an analysis of the constrained variety of poses and garments in the generated results, which can be found in Appendix~\ref{sec:moreresult}. To amplify the generative capabilities of the model and promote diversity in poses and garments, we anticipate that incorporating data augmentation techniques and more varied datasets could serve as the subsequent steps.

\noindent
\textbf{Conclusion.} This work proposes an end-to-end training pipeline with the goal of orchestrating multi-source human-centric datasets with various distributions and scales into a full-body image space and achieving high-resolution human synthesis. The Multi-source Spatial Transformer, in particular, copes with articulated human structure, while the Continuous GAN module enables producing images at different resolutions. 
UnitedHuman breaks the technical barrier of being unable to generate high-fidelity human bodies in the absence of adequate HD full-body images and opens up new research directions to accelerate the process of human-body generation.

\noindent
\textbf{Acknowledgements.} This study is supported by NTU NAP, MOE AcRF Tier 1 (2021-T1-001-088), and under the RIE2020 Industry Alignment Fund – Industry Collaboration Projects (IAF-ICP) Funding Initiative, as well as cash and in-kind contribution from the industry partner(s).

{\small
\bibliographystyle{ieee_fullname}
\bibliography{references}
}

\clearpage
\appendix

\noindent
\textbf{\LARGE Appendix}

\vspace{6mm}

In this appendix, we first show how we construct the synthetic hand dataset~(Sec.~\ref{hand}). Then we provide the experimental settings for the Multi-source Spatial Transformer~(MST), Continuous GAN, as well as the overall training configuration~(Sec.~\ref{exp}). Additionally, we provide more qualitative results of ablation studies~(Sec.~\ref{qualitative}). We also conduct a user study to evaluate the performance of our method and the other three SOTA methods~(Sec.~\ref{userstudy}). Finally, an analysis in terms of the poses and garments diversity in the generated images is given in Sec.~\ref{sec:moreresult}.

\section{Synthetic Hand Dataset}\label{hand}
\begin{figure}[h]
    \centering
    \includegraphics[width=\linewidth]{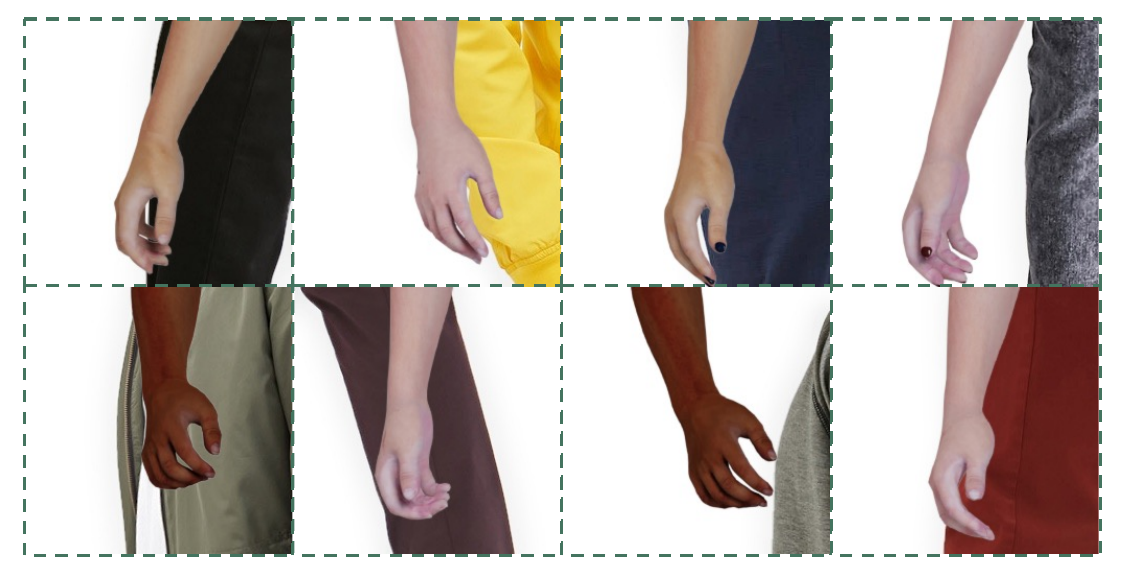}
    \captionsetup{skip=0pt}
    \caption{Synthetic Hand Dataset. We generate the training images by compositing the hand images rendered from the DART dataset with image patches from DeepFashion-HR datasets.}
    \label{fig:hand}
\end{figure}

Due to the lack of high-quality hand datasets for generative tasks, we construct a new dataset based on the DART dataset and SMPL-X model. Specifically, we first randomly sample $7,000$ images from the full-body SHHQ dataset for SMPL-X estimation, where the estimated parameters include hand pose and position. Then, we apply the sampled DART texture to the palm and arm, remove the mesh except for the arm, and introduce a photo studio HDR as the ambient light. The Eevee renderer~\cite{blender} is used to render the hand images in the full-body image space with the average camera $\alpha$ of SHHQ. As shown in Fig.~\ref{fig:hand}, we finally composite the synthesized hand with an image patch from the DeepFashion-HR dataset.

\section{Experiment settings}\label{exp}
In this section, we provide the experiment settings of our framework. 
Multi-source Spatial Transformer aims to transform the partial-body image into the full-body image space by estimating the camera parameters of SMPL. We divide the objective function of SMPLify-p into three parts:
\begin{equation}
    \begin{split}
    E(\alpha_{opt}) 
    &= \lambda_{v}L_{vis} + \lambda_{i}L_{invis} + \lambda_{r}L_{reg}\\
    \end{split}
\end{equation}
where $\lambda_{v}$, $\lambda_{i}$, $\lambda_{r}$ is $1.0$, $10.0$ and $10.0$ respectively. The loss $L_{invis}$ is based on the variational autoencoder~(VAE) trained on the full-body SHHQ dataset. In particular, we follow the architecture of VPoser~\cite{SMPL-X:2019} that contains a specific decoder for continuous rotation representation.  The loss function of VAE consists of reconstruction loss on vertices and KL divergence. We use the pose parameters estimated by PARE~\cite{pare} as ground truth for training.

As for the Continuous GAN, we follow the training setting of the original StyleGAN3-T except for the R1 gamma, which is set to $2$ in our experiments. Both \textit{Stage 1} and \textit{Stage 2} training use a batch size of $32$ and are trained on $8$ GPUs.
Training configurations for all the compared methods can be found in Tab.~\ref{tab:trainingparams}.

\begin{table}[tbp]
    \begin{center}
    \small
    \resizebox{\linewidth}{!}{
        \begin{tabular}{ccccc}
        \toprule 
         & SG-Human & InsetGAN$^*$ & AnyRes & Ours \\
        \midrule
        Batch Size & 32 & / & 16 & 32\\
        G\_Learning rate & 0.002 & / & 0.0025 & 0.0025\\
        D\_Learning rate & 0.002 & / & 0.002 & 0.002 \\
        Total \#Params. (M) & 30 / 29 & / & 29 / 31 & 28 / 55 \\ 
        Train (GPU Days) & 7 & / & 23 & 43 \\ 
        Test (GPU Sec.) & 0.48 & 200 & 2.6 & 2.8 \\
        \bottomrule 
        \end{tabular}       
    }
    \end{center}
    \setlength{\abovecaptionskip}{-5pt} 
    \setlength{\belowcaptionskip}{-10pt} 
    \caption{Training details of experiments.
    InsetGAN$^*$ is an optimization-based method and generates results without retraining. }
    \label{tab:trainingparams}
\end{table}

\section{Additional qualitative results}\label{qualitative}

\begin{figure}
    \centering
    \includegraphics[width=1.05\linewidth]{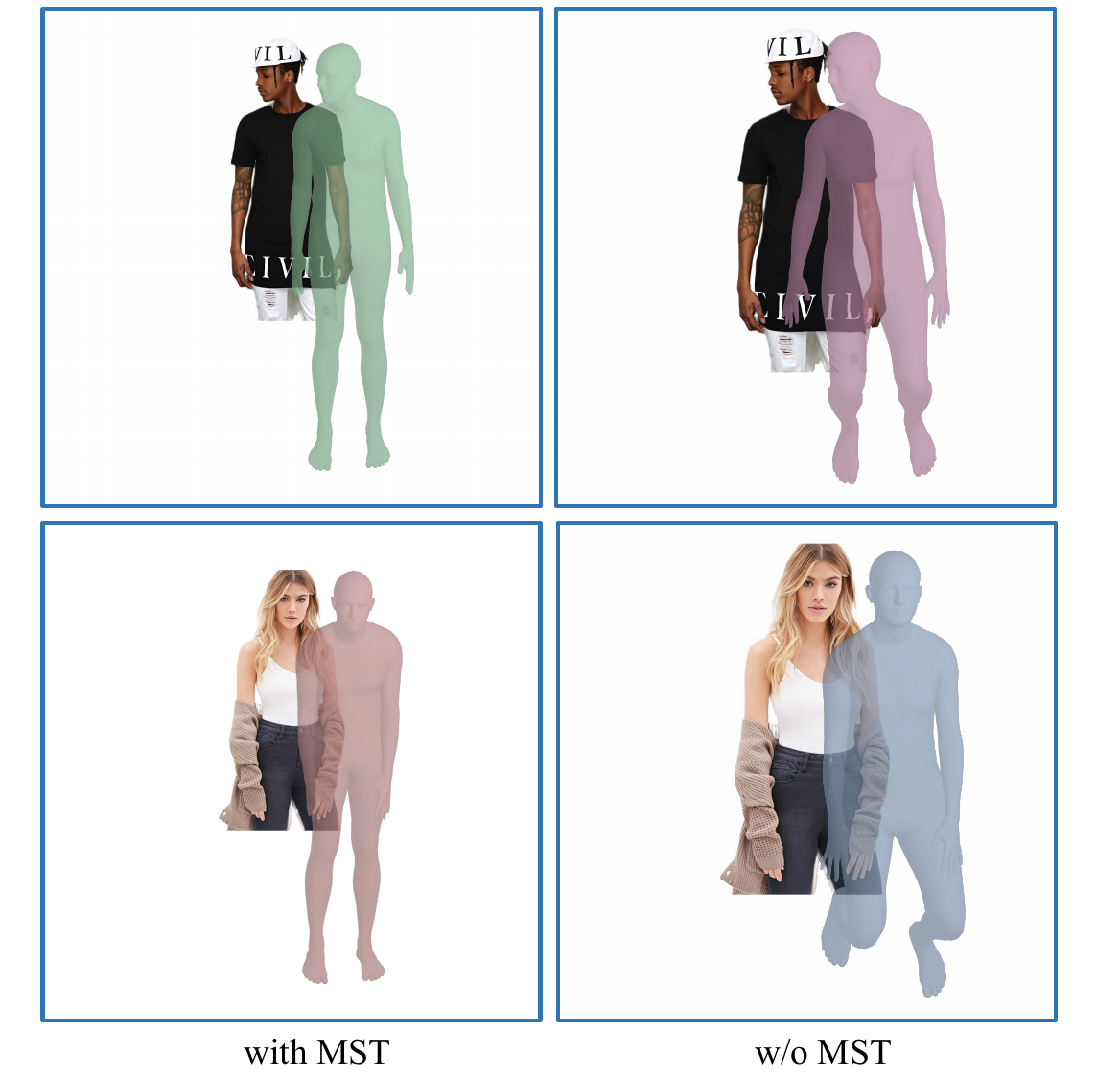}
    \captionsetup{skip=0pt}
    \caption{Visualization results of ablation on Multi-source Spatial Transformer.}
    \label{fig:mst}
\end{figure}

\noindent\textbf{Ablation on multi-source spatial transformer.} Fig.~\ref{fig:mst} displays the partial-body images that were transformed into the full-body image space, with and without Multi-source Spatial Transformer (MST). In the right column, the reconstructed parametric mesh does not match the reference image when MST is not used. In contrast, the spatial distribution of images transformed by MST is closer to that of the full-body dataset.

\begin{figure*}[ht]
    \centering\includegraphics[width=0.9\textwidth]
    {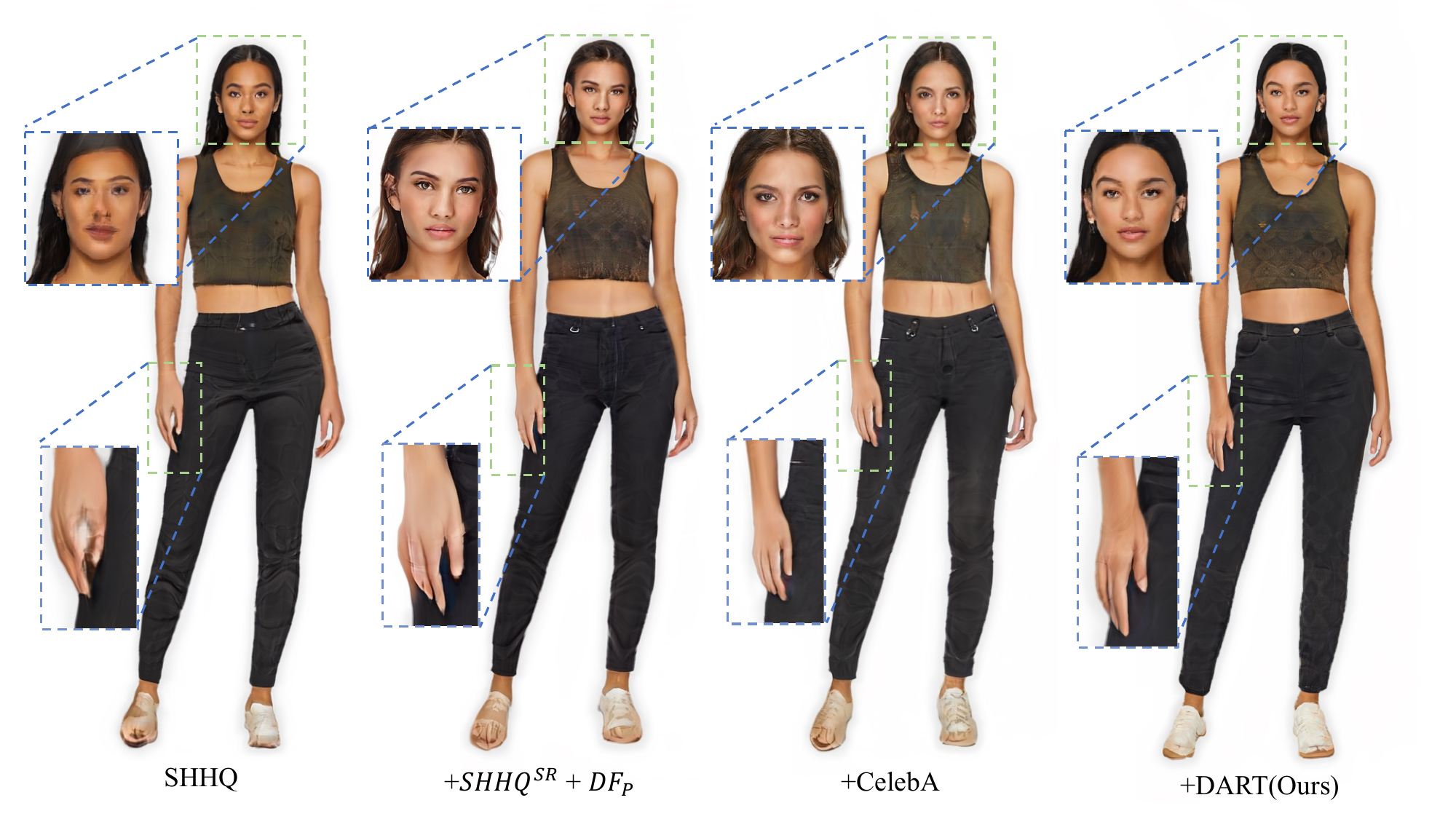}
    \captionsetup{skip=0pt}
    \caption{Visualization of ablation on datasets. From left to right, the visualizations are the generated results after continuously adding datasets from different sources to the model.}
    \vspace{-2mm}
    \label{fig:vis_data_ablation}
\end{figure*}

\noindent\textbf{Ablation on multi-source datasets.} Fig. \ref{fig:vis_data_ablation} demonstrates the improvement of local details by using more multi-source partial datasets. The full-body images are 1024px, while the local patches are cropped from the 2048px images. When only trained on SHHQ, our model is capable of generating coherent full-body images with a resolution of 1024 pixels. However, at higher scales, the generated patches are blurry and have artifacts. With SHHQ\textsuperscript{SR} and $\text{DF}_p$ datasets, both the face and hand patches are clearer than before. By adding the face dataset CelebA, more details such as the illumination on the face are captured. Finally, with the addition of a constructed synthetic hand dataset, the details of hand patches are more accurate.

\begin{figure}
    \centering
    \includegraphics[width=\linewidth]{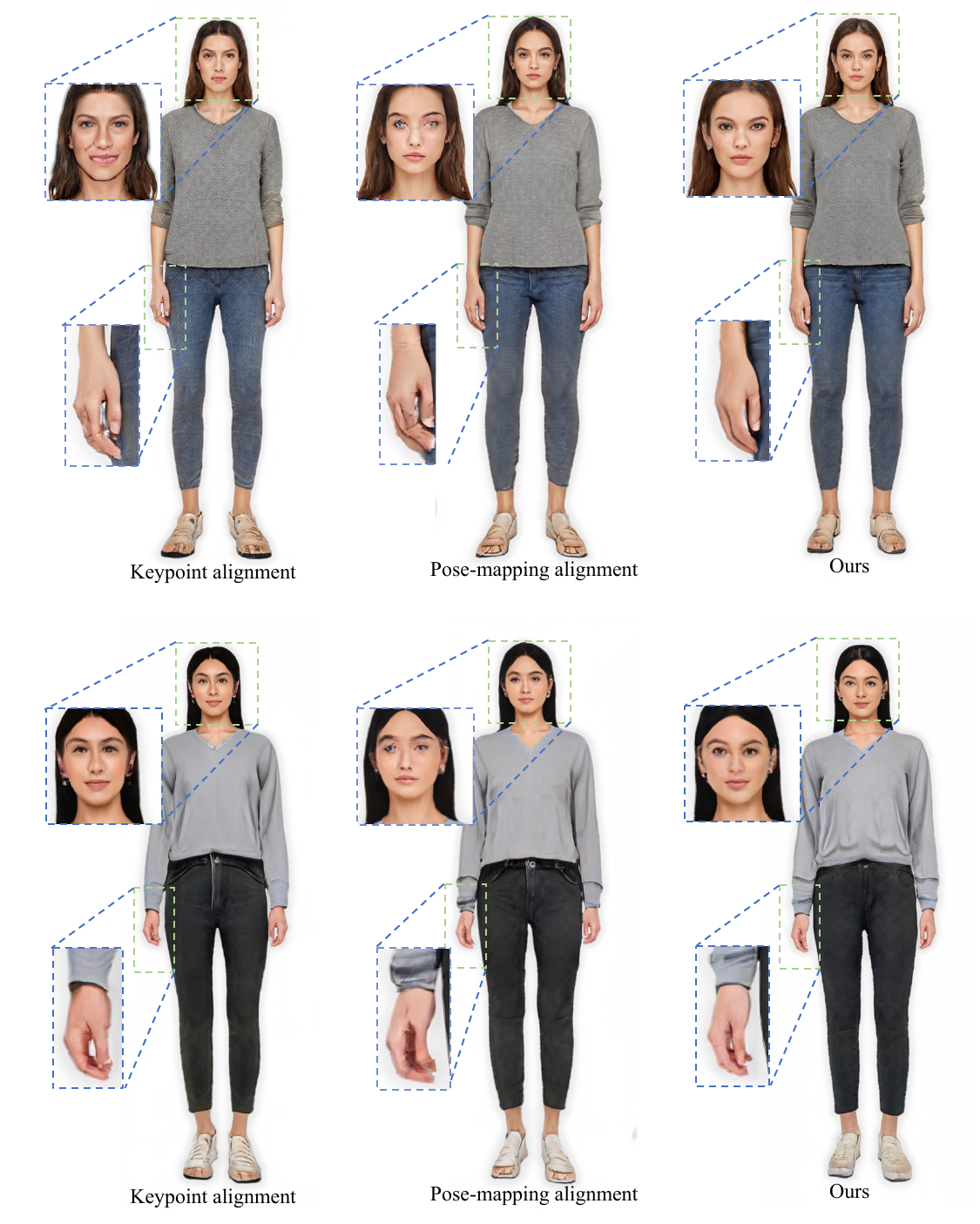}
    \captionsetup{skip=0pt}
    \caption{Visualization of ablation on alignment strategy.}
    \label{fig:vis_alignment_mer}
\end{figure}

\noindent\textbf{Ablation on alignment strategy.} We provide visual comparisons~(Fig.~\ref{fig:vis_alignment_mer}) of the ablation study conducted on different human alignment strategies. The top row of the figure contains the images generated by the mean latent of each model, and the images in the bottom row are generated by a random latent. It can be observed from the figure that the visual outcomes are consistent with the kFIDs reported in the paper. Aligning humans using only 2D keypoints yields superior face and hand details compared to using an auxiliary ``pose-mapping'' MLP network. Nonetheless, our proposed approach outperforms these two methods in producing finer details at a higher resolution. 

\begin{figure}[pb]
    \vspace{-2mm}
    \centering
    \includegraphics[width=1\linewidth]{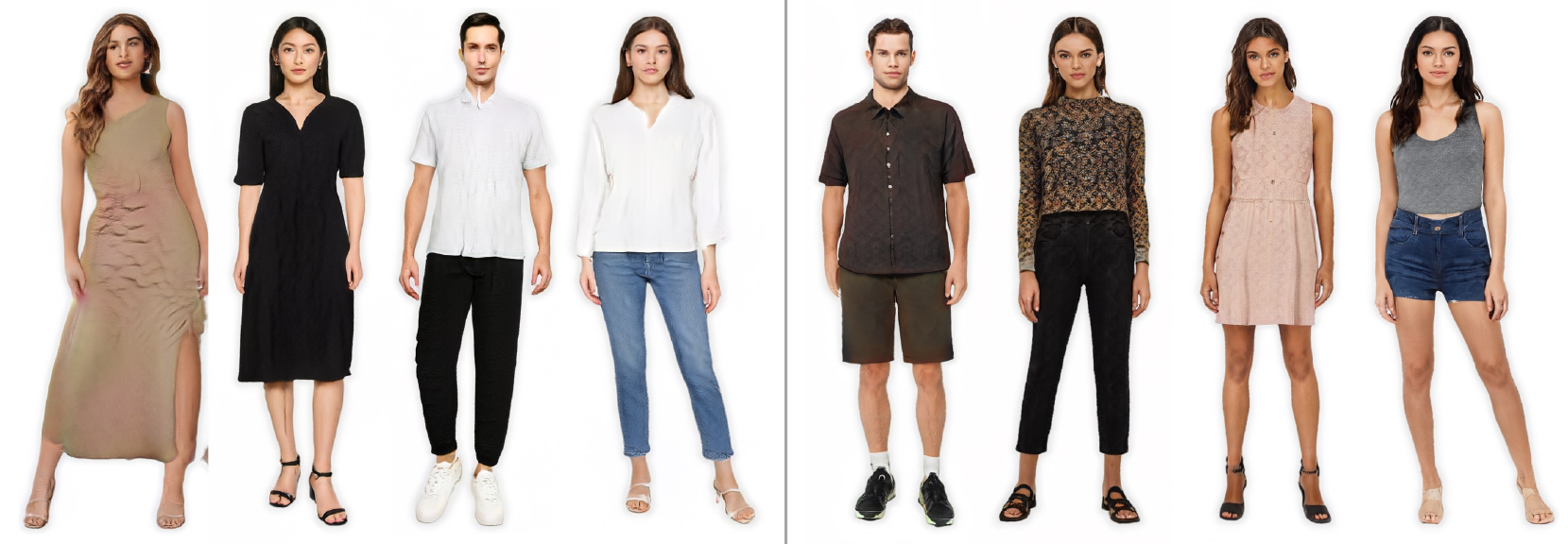}
    \setlength{\abovecaptionskip}{-5pt} 
    \setlength{\belowcaptionskip}{-10pt} 
    \caption{Left: StyleGAN-Human~(10K).  Right: Ours~(10K). Experiments show that the constrained diversity of garments and poses is attributed to the training data.
    }
    \label{fig:sghuman_1w_sample}
\end{figure}

\section{User Study}\label{userstudy}
\begin{figure}[ht]
    \centering
    \includegraphics[width=\linewidth]{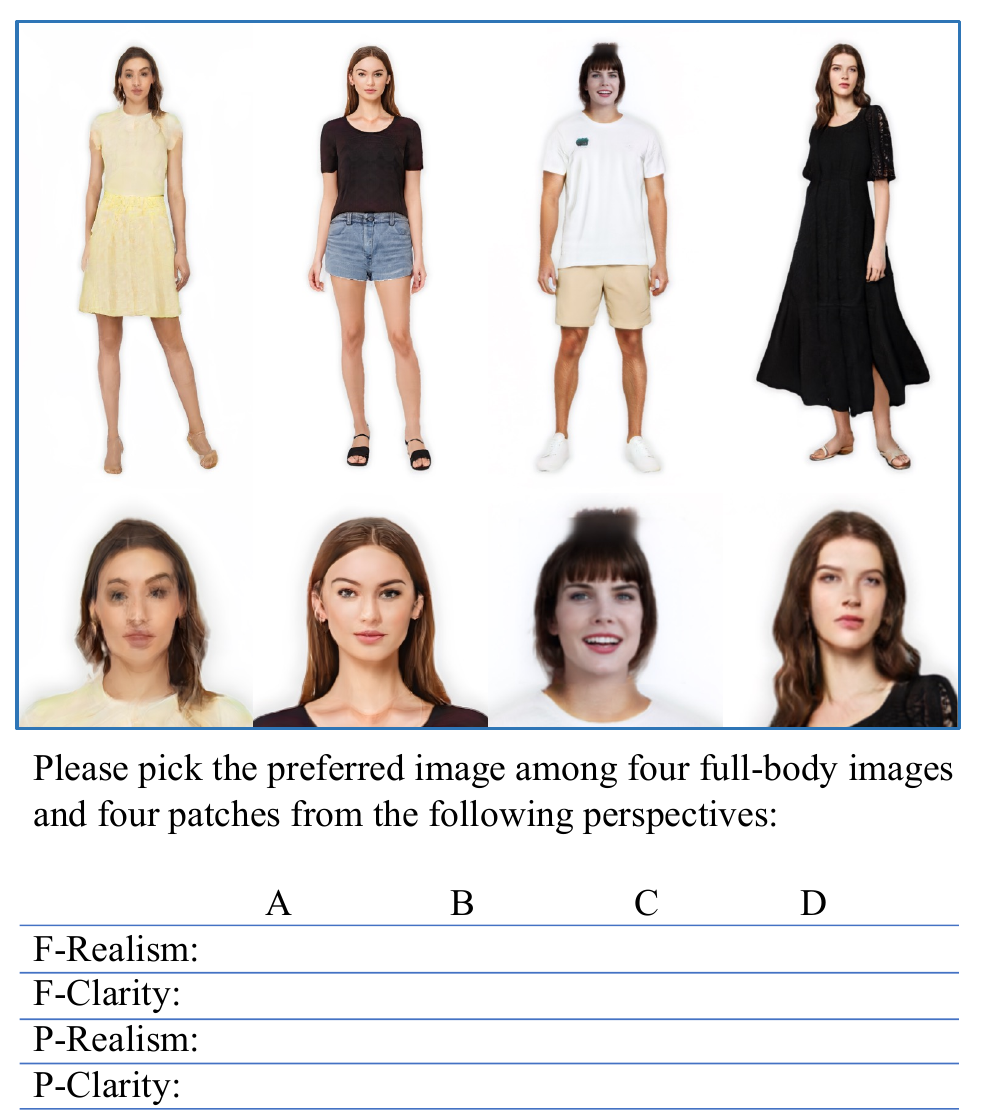}
    \captionsetup{skip=0pt}
    \caption{This is an example of the questionnaire used in our user study.}
    \label{fig:userstudy_questionnaire}
\end{figure}

We conduct a user study to evaluate the clarity and realism of the results generated by our method in comparison to three SOTA methods: StyleGAN-Human~\cite{styleganhuman}, InsetGAN~\cite{insetgan} and AnyRes~\cite{anyres}. Our user study involves a total of $12$ volunteers. We randomly selected $10$ high-resolution ($2048$px) images from each model and extracted the face or hand regions from those images as well. During the user study, the participants were presented with four images at a time, along with the corresponding cropped patches. Every participant was asked to evaluate the authenticity and sharpness of both the full images and their cropped parts. One of the examples that appeared in our questionnaire is displayed in Fig.~\ref{fig:userstudy_questionnaire}. Based on global realism, global clarity, and local realism, the images generated by our method received over $82\%$ of the votes, while the clarity of the local patches was slightly lacking, receiving a score of $78\%$. Nonetheless, compared to the other three SOTA methods, our approach is still far ahead in visual quality. More detailed results are shown in Tab.~\ref{tab:userstudyv1}.

\begin{table}[htbp]
    \centering
    \small
    \begin{tabular}{l|c|c|c|c}
    \toprule 
    & SG-Human & InsetGAN & AnyRes & Ours \\
    \midrule
     F-Realism & $13.33$ & $0.83$ & $0.83$ & $85$\\ 
     F-Clarity & $8.33$ & $3.33$ & $0$ & $88.33$ \\
    P-Realism & $6.66$& $9.17$ & $1.67$ & $82.5$ \\
    P-Clarity & $2.5$ & $17.5$ & $1.67$ & $78.33$ \\
    \bottomrule 
    \end{tabular}
    \setlength{\belowcaptionskip}{-10pt} 
    \caption{Voting scores for images from both SOTAs and our model, which are evaluated based on four criteria: the full-image's realism and clarity, as well as the cropped regions' realism and clarity. The table uses ``F-" and ``P-" to denote the full images and cropped patches, respectively. All the numbers in the table are presented in percentages ($\%$). Higher values indicate better realism or clarity.}
    \label{tab:userstudyv1}
\end{table}

\section{More results}
\label{sec:moreresult}
The showcased human images demonstrate neutral poses and relatively less diverse garments. In response to this situation, we conducted an additional experiment using $10K$ SHHQ images on StyleGAN-Human to confirm that the limited diversity of pose and clothing is mainly due to the training dataset~(see Fig.~\ref{fig:sghuman_1w_sample}). Specifically, the majority of the poses and garments in our model are sourced from the $10K$ SHHQ and $10K$ DeepFashion images, while StyleGAN-Human utilizes re-balancing techniques and a more extensive dataset of $230K$ images.

\end{document}